\documentclass{article}

\usepackage{microtype}
\usepackage{graphicx}
\usepackage{subcaption}
\usepackage{booktabs} 
\usepackage{tabularx}
\usepackage{array}
\usepackage{tikz}
\usetikzlibrary{shapes.geometric, arrows.meta, positioning, calc}
\usepackage{pifont}

\usepackage{hyperref}
\usepackage{float}
\usepackage{dblfloatfix}

\usepackage[preprint]{icml2026}

\usepackage{amsmath}
\usepackage{amssymb}
\usepackage{mathtools}
\usepackage{amsthm}

\usepackage[capitalize,noabbrev]{cleveref}

\theoremstyle{plain}

\theoremstyle{definition}

\theoremstyle{remark}

\usepackage[textsize=tiny]{todonotes}

\icmltitlerunning{Location-Aware Pretraining for Medical Difference Visual Question Answering}

\begin{document}

\twocolumn[
  \icmltitle{Location-Aware Pretraining for Medical Difference Visual Question Answering}



  \icmlsetsymbol{equal}{*}

  \begin{icmlauthorlist}
    \icmlauthor{Denis Musinguzi}{yyy}
    \icmlauthor{Caren Han}{sch}
    \icmlauthor{Prasenjit Mitra}{yyy}
  \end{icmlauthorlist}

  \icmlaffiliation{yyy}{Department of Electrical and Computer Engineering, Carnegie Mellon University, Kigali, Rwanda}
  \icmlaffiliation{sch}{University of Melbourne, Melbourne, Australia}

  \icmlcorrespondingauthor{Denis Musinguzi}{dmusingu@andrew.cmu.edu}
  \icmlcorrespondingauthor{Prasenjit Mitra}{prasenjm@andrew.cmu.edu}

  \icmlkeywords{Machine Learning, ICML}

  \vskip 0.3in
]



\printAffiliationsAndNotice{}  

\begin{abstract}
Differential medical VQA models compare multiple images to identify clinically meaningful changes, relying on vision encoders to capture fine-grained visual differences in ways that mirror radiologists’ comparative diagnostic workflows. However, vision encoders trained with standard contrastive or classification objectives often fail to capture the subtle variations required to distinguish true disease progression from acquisition-related variability. To address this limitation, we introduce a location-aware pretraining framework that incorporates automatic referring expressions (AREF), grounded captioning (GCAP), and conditional automatic referring expressions (CAREF). These tasks encourage the learning of fine-grained, spatially grounded visual representations. Integrated with a language model, our approach achieves state-of-the-art performance on medical difference VQA, effectively identifying and reasoning about clinically relevant changes in chest X-ray images.
\end{abstract}

\begin{figure*}
\centering
\begin{tikzpicture}
    [
    node distance=2cm,
    scale=0.5,           
    transform shape,
    block/.style={
        rectangle, rounded corners=5pt, draw, thick,
        minimum height=1.2cm, minimum width=1.2cm,
        fill=gray!10, font=\normalsize\sffamily
    },
    trapezoid/.style={
        trapezium,
        trapezium left angle=75,
        trapezium right angle=75,
        rotate=-90,
        draw=gray!80,
        top color=white,
        bottom color=red!10, 
        very thick,
        inner sep=12pt,
        font=\small\sffamily\bfseries,
        text centered,
        minimum height=3.5cm 
    },
    cube/.style={
    rectangle, draw, thick,
    minimum size=1.2cm, font=\tiny\sffamily, align=center
    },
    textenc/.style={
        trapezium, shape border rotate=-90,
        draw=red!70!black, fill=red!15, thick,
        minimum width=2.5cm, minimum height=1.5cm,
        font=\small\sffamily\bfseries, align=center
    },
    input_text/.style={
        rectangle, rounded corners=12pt,
        draw=orange!80, fill=orange!10, thick,
        text width=5cm, minimum height=2.5cm,
        font=\small\sffamily, align=left, inner sep=10pt
    },
    decoder/.style={
        rectangle, rounded corners=15pt,
        draw=green!50!black, fill=green!20, thick,
        minimum width=3.5cm, minimum height=7cm,
        font=\sffamily\bfseries, align=center
    },
    vit/.style={
        trapezium,
        shape border rotate=270, 
        trapezium left angle=75,
        trapezium right angle=75,
        draw=green!60!black,
        fill=green!15,
        thick,
        minimum height=1.5cm, 
        minimum width=2.5cm,
        font=\sffamily\bfseries,
        align=center
    },
    >={Stealth[length=3mm]},
    thick
    ]

    \node[inner sep=0](img1){\includegraphics[width=2.5cm]{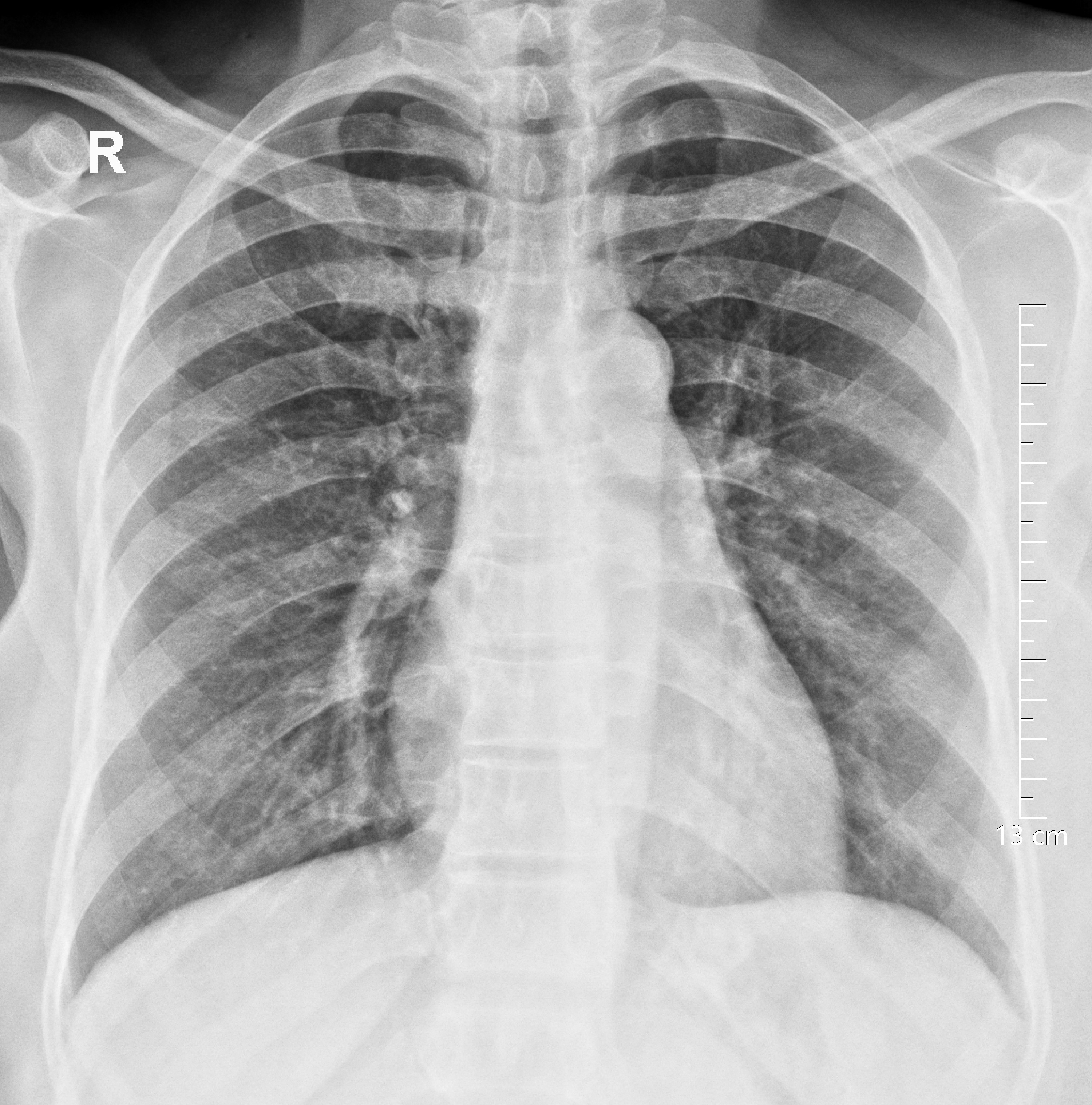}};
    \node[above=0.1cm of img1, font=\footnotesize\sffamily]{Prior image};

    \node[vit, right=of img1] (vit1) {ViT};

    \node[block, draw=brown!60, fill=blue!10, right=of vit1, sharp corners] (feat1) {\rotatebox{90}{Prior image features}};

    \node[block, right=of feat1] (proj1) {Projection Layer};

    \node (plus1) [draw, circle, inner sep=2pt, right=of proj1] {$+$};

    \node[inner sep=0, below=1.4cm of img1](img2){\includegraphics[width=2.5cm]{images/59.jpg}};
    \node[above=0.1cm of img2, font=\footnotesize\sffamily]{Current image};

    \node[vit, right=of img2] (vit2) {ViT};

    \node[block, draw=brown!80, fill=brown!20, right=of vit2, sharp corners] (feat2) {\rotatebox{90}{Current image features}};

    \node[block, right=of feat2] (proj2) {Projection layer};
    \node (plus2) [draw, circle, inner sep=2pt, right=of proj2] {$+$};

    \node[block, above=1cm of plus1] (temp1) {Temporal encoding};
    
    \node[block, below=1cm of plus2] (temp2) {Temporal encoding};

    \node[block, minimum width=1cm, minimum height=4.0cm, right=of plus1, sharp corners, draw=brown!60, fill=blue!10] (imgfeat1) {};

    \node[block, minimum width=1cm, minimum height=4.0cm, below=0cm of imgfeat1, sharp corners, draw=brown!80, fill=brown!20] (imgfeat2) {};

    \node[rectangle, rounded corners=5pt, draw=orange!80, fill=orange!10, text width=5cm, below=2.5cm of img2, xshift=2cm, minimum height=4.3cm] (input_text) {Increased right pleural effusion Left lower lobe  pneumonia is improving.};

    \node[vit, right=5cm of input_text, draw=red!70!black,
      fill=red!15, minimum width=3cm, minimum height=4cm] (text_encoder) { Text\\ Encoder};

    \node[block, minimum width=1cm, minimum height=4.0cm, below=2.0cm of imgfeat2, sharp corners] (text_features) {};

    \node[decoder, right=5cm of temp2, yshift=1cm] (text_decoder) {Transformer decoder};
      
    \draw[dashed, gray] (vit1) -- (vit2) node[midway, right, font=\tiny\sffamily] {Shared weights};
    
    \draw[->](img1)--(vit1);
    \draw[->](img2)--(vit2);
    \draw[->](vit1)--(feat1);
    \draw[->](vit2)--(feat2);
    \draw[->](feat1)--(proj1);
    \draw[->](feat2)--(proj2);
    \draw[->](proj1)--(plus1);
    \draw[->](proj2)--(plus2);
    \draw[->](temp1)--(plus1);
    \draw[->](temp2)--(plus2);
    \draw[->](plus1)--(imgfeat1);
    \draw[->](plus2)--(imgfeat2);
    \draw[->](input_text)--(text_encoder);
    \draw[->](text_encoder)--(text_features);
    \draw[->] (text_features.east) -- ++(1cm,0) |- (text_decoder.west);
    \draw[->] (imgfeat2.north east) -- ++(1cm, 0) |- (text_decoder.west);
    
\end{tikzpicture}
\caption{\textbf{Overview of medical difference model architecture}. The model consists of a frozen pretrained vision encoder, a vision adapter, and a transformer decoder.}
\label{fig:med-diff-architecture}
\end{figure*}

\section{Introduction}
Visual Question Answering (VQA) aims to answer questions about visual content by jointly interpreting image and text inputs. Prior work has shown strong performance on VQA tasks involving a single medical image \cite{sellergren2025medgemmatechnicalreport, li2023llavamedtraininglargelanguageandvision, zhang2024pmcvqavisualinstructiontuning, zhang2025biomedclipmultimodalbiomedicalfoundation}. However, this single-image formulation is poorly aligned with clinical diagnostic practice, where key decisions often depend on identifying subtle changes across multiple imaging studies rather than interpreting an image in isolation.

In many clinical settings, radiologists perform longitudinal comparisons of medical images from the same patient to assess disease progression and treatment response. For example, during tuberculosis (TB) treatment, chest X-ray images are acquired over time to determine whether thoracic abnormalities are improving, worsening, or remaining stable—differences that may be visually subtle yet clinically decisive. Such comparative reasoning cannot be supported by single-image analysis, hence the need for multi-image visual understanding. Medical difference VQA \cite{Hu_2023} addresses this gap by extending VQA to a multi-image setting in which models must reason explicitly about changes between images to answer clinically meaningful questions.

Medical difference VQA is substantially more challenging than conventional VQA because, even for the same patient, disease-related changes may be visually comparable in magnitude to variations caused by image acquisition, such as differences in orientation, scale, field of view, or nonrigid deformation across scans \cite{Hu_2023}. As a result, clinically relevant changes can be extremely difficult to detect. This challenge is particularly pronounced in chest X-ray images, where pathologies often manifest in subtle and localized ways. Consequently, models must acquire a fine-grained understanding of pathological visual patterns in order to detect disease-related changes while disregarding irrelevant variations \cite{yao2022image}.

To tackle medical difference VQA, existing approaches typically adopt architectures composed of a shared vision encoder, a language model, and a vision adapter \cite{cho2024pretraining, bannur2023learning, yang2025medical, chen2025coca} (see \cref{fig:med-diff-architecture}). In this setup, multiple images are processed independently by the same image encoder. To encode temporal ordering, image index embeddings are added to the extracted visual features \cite{bannur2023learning} before they are projected by the vision adapter into a shared multimodal embedding space.

Despite their effectiveness, these architectures commonly rely on vision encoders pretrained on natural images using objectives such as ImageNet classification or CLIP-style contrastive learning \cite{radford2021learning, jia2021scaling}. While such pretraining yields strong global semantic representations, it often fails to provide the fine-grained visual grounding \cite{tong2024eyes} required for medical difference VQA, where subtle regional changes are critical. Moreover, because standard training pipelines typically freeze the vision encoder, the resulting visual representations remain suboptimal for medical imaging tasks, further limiting the model’s ability to capture clinically meaningful differences.

In this work, we address these limitations by introducing a domain-adaptive pretraining stage in which the vision encoder is optimized using location-aware objectives that promote a fine-grained understanding of anatomical structures, while remaining aligned with global image semantics. Building on the multi-task generative framework of \cite{beyer2023study}, we combine \textbf{global captioning} with three complementary forms of region-level supervision (see \cref{fig:locca-architecture}): (i) \textbf{automatic referring expressions (AREF)}, which requires predicting bounding box coordinates from automatically generated captions describing specific image regions; (ii) \textbf{conditional automatic referring expressions (CAREF)}, which extends AREF by conditioning box prediction on anatomical region names in addition to generated captions; and (iii) \textbf{grounded captioning (GCAP)}, which requires predicting both region-level bounding boxes and their corresponding captions from the image. By jointly optimizing global image captioning and localized grounding objectives, this pretraining scheme aligns coarse semantic understanding with fine-grained, spatially grounded representations, enabling the encoder to better capture subtle anatomical details and spatial relationships critical for medical difference reasoning.

Following this location-aware pretraining stage, we freeze the vision encoder and jointly fine-tune the multimodal projection layer and language model for the medical difference VQA task.
Our contributions can be summarized as follows:
\begin{itemize}
\item We introduce a multimodal pretraining framework that augments standard auto-regressive training of medical vision--language models with location-aware objectives, including grounded captioning, automatic referring expressions, and conditional automatic referring expressions.
\item We perform comprehensive ablation studies on the proposed location-aware tasks, showing that each objective contributes to improved visual representation learning, with their combined use yielding the strongest performance.
\item We demonstrate that this grounding-centric pretraining strategy substantially improves the vision encoder’s adaptation to the medical domain, resulting in significant gains on the challenging medical difference VQA task.
\end{itemize}

\section{Related Work}
\label{sec:related work}
Advances in vision–language models (VLMs), such as CLIP \cite{radford2021learning}, have demonstrated strong performance in global image–text alignment. However, these models primarily capture coarse, image-level semantics and often struggle to represent fine-grained, region-specific visual cues that are critical in medical imaging, where clinically relevant findings occupy very small regions. This limitation has prompted growing interest in pretraining strategies that explicitly model localized visual semantics.


\subsection{Local Contrastive Learning} 
Recent contrastive learning frameworks aim to jointly model global and local visual representations. For example, GLoRIA \cite{huang2021gloria} uses an attention-based mechanism to align image regions with individual words in radiology reports. However, individual words often lack sufficient semantic context to describe complex visual patterns, and learning such alignments without explicit supervision can result in noisy or unreliable associations.

Other approaches, such as VLMAE \cite{he2022vlmae}, combine visual generative learning with contrastive objectives. By reconstructing masked image patches using both regional textual descriptions and visible patches, these methods encourage the learning of local visual features while maintaining global image–text alignment. Nevertheless, they rely on implicit and potentially ambiguous alignment signals. In contrast, we exploit the explicit region–phrase annotations provided by the Chest ImaGenome dataset \cite{wu2021chestimagenome} to directly supervise the learning process through location-aware tasks.



\subsection{Multi-task Generative Pretraining} 
Recent studies \cite{tschannen2023image, fini2025multimodal} have shown that generative captioning objectives can outperform contrastive learning for vision–language tasks. Building on this insight, the flexibility of transformer architectures has enabled the unification of multiple tasks through task-specific prompting \cite{beyer2023study, li2022grounded, zhang2022glipv2}, allowing models to progress beyond coarse global descriptions toward fine-grained, region-specific understanding. By incorporating location-aware tasks such as grounded captioning and referring expression comprehension, models are explicitly trained to localize visual features \cite{wan2024locca}. We exploit this capability to address the high granularity demands of medical image understanding, where precise localization of abnormalities is essential.

\subsection{Longitudinal modeling of medical images}

Approaches to medical difference VQA can be broadly categorized into three paradigms: \textbf{feature concatenation, explicit difference modeling}, and \textbf{intermediate reasoning.} Concatenation-based methods \cite{cho2024pretraining, yao2022image} process multiple image streams using a shared visual encoder and subsequently combine their representations. For example, PLURAL \cite{cho2024pretraining} adopts a multi-stage transfer learning strategy from natural images and concatenates features extracted from prior and current scans. While effective at capturing global contextual changes, these approaches primarily rely on coarse feature alignment and often fail to represent the fine-grained, localized differences that are critical for accurate medical comparison.

Explicit difference modeling methods, such as ReAL \cite{10.1007/978-3-031-72086-4_61, hu2023expert}, directly compute visual changes by generating pixel-level residual images, which are then processed by a separate visual encoder. Although conceptually appealing, this approach is highly sensitive to spatial misalignment between scans. Without careful image registration, simple pixel-wise subtraction can amplify noise and introduce spurious differences, ultimately degrading the robustness of the VQA pipeline \cite{10.1007/978-3-031-72086-4_61}.

Intermediate reasoning approaches, including RG‑AG \cite{serra2025grounding}, employ a two-stage pipeline in which a descriptive radiology report is first generated and then used as contextual input to answer downstream questions. While this strategy effectively leverages the reasoning capabilities of large language models, it introduces an information bottleneck: visual details not captured in the intermediate textual description are irreversibly lost. In contrast, our method avoids both fragile pixel-level differencing and lossy textual intermediates by introducing a domain-adaptive pretraining framework that directly enforces fine-grained, location-aware visual understanding—an essential requirement for reliable medical difference reasoning.

\begin{figure*}
\begin{tikzpicture}[
    node distance=2.5cm,
    scale=0.5,
    block/.style={
        rectangle,
        rounded corners=8pt,
        draw=blue!70,
        fill=blue!15,
        thick,
        align=center,
        inner sep=10pt,
        text=black,
        font=\small\sffamily\bfseries,
        minimum height=2.5cm
    },
    textbox/.style={
        rectangle,
        rounded corners=5pt,
        thick,
        fill=white,
        font=\scriptsize\sffamily
    },
    examplebox/.style={
        rectangle,
        rounded corners=5pt,
        thick,
        fill=white,
        font=\scriptsize\sffamily,
        minimum height=1.5cm,
    },
    >={Stealth[length=2.5mm]},
    thick
]

\node[block, minimum width=4cm, minimum height=2cm] (decoder) {Transformer\\Decoder};
\node[block, minimum width=4.0cm, minimum height=2cm, left=4cm of decoder] (vit) {Vision\\Transformer};

\node[draw, thick, inner sep=0, below=1.0cm of vit, name=img] {\includegraphics[width=2cm]{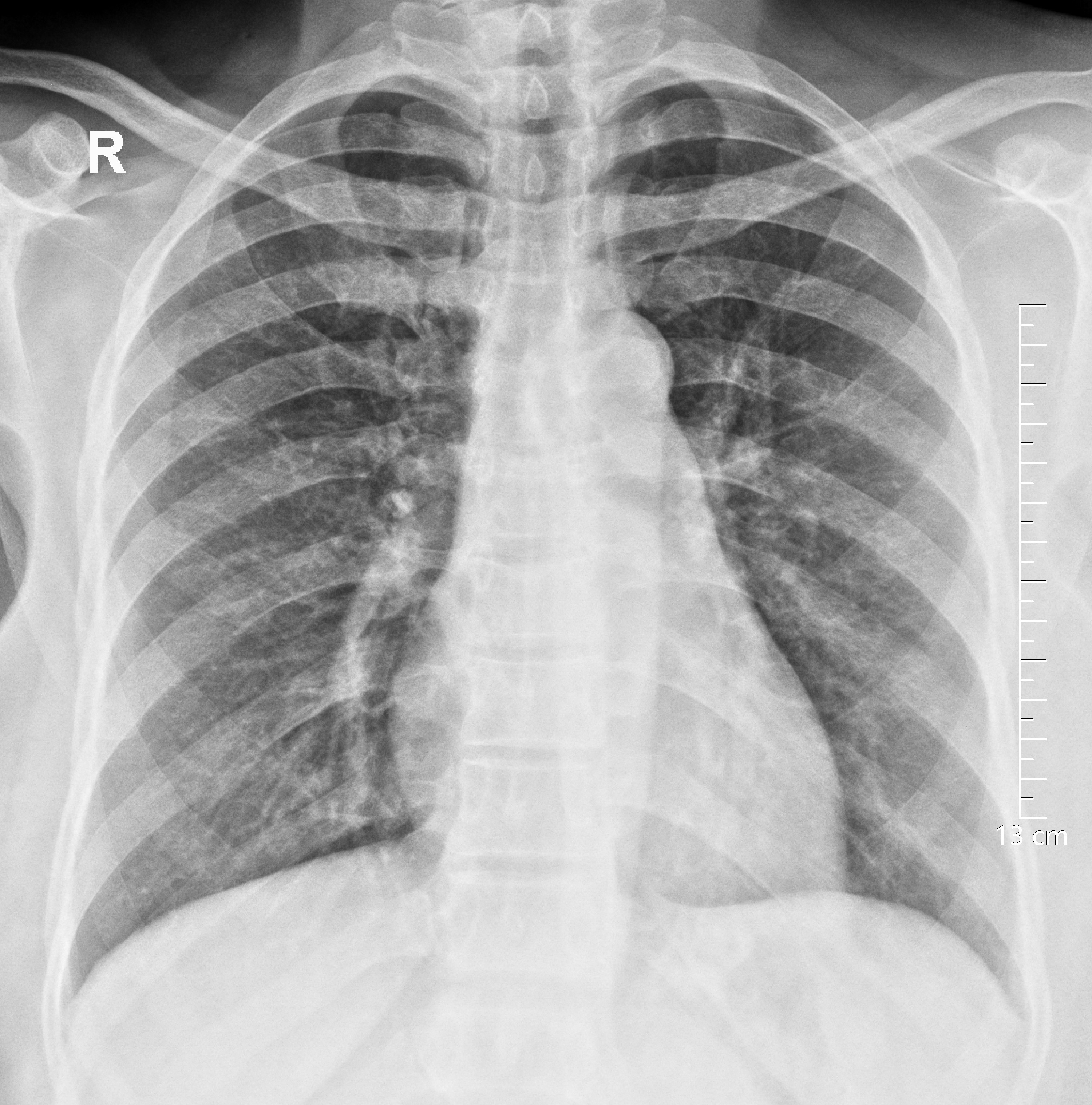}};

\draw[->] (img) -- (vit);

\draw[->] (vit) -- (decoder) node[midway, above, font=\small\sffamily, text width=2.5cm, align=center] {Cross Attention};

\def\topoffset{4.5}
\def\topspread{2.6} 

\coordinate (T1) at ($ (decoder.north west)!0.15!(decoder.north east) $);
\coordinate (T2) at ($ (decoder.north west)!0.38!(decoder.north east) $);
\coordinate (T3) at ($ (decoder.north west)!0.62!(decoder.north east) $);
\coordinate (T4) at ($ (decoder.north west)!0.85!(decoder.north east) $);

\node[examplebox, draw=black, text width=2.5cm, align=left, inner sep=8pt] 
    (top1) at ($ (decoder.north) + (-4.2*\topspread, \topoffset) $) 
    {There is no pulmonary edema.};

\node[examplebox, draw=green!70!black, text width=2.5cm, align=left, inner sep=8pt, right=3.5mm of top1] 
    (top2) {There is no pulmonary edema. \relax{[61, 66, 95, 110]}};

\node[examplebox, draw=blue!70, text width=2.5cm, align=left, inner sep=8pt, right=3.5mm of top2] 
    (top3) {{\relax{[61, 66, 95, 110]}} There is no pulmonary edema.};

\node[examplebox, draw=red!70!black, text width=2.5cm, align=left, inner sep=8pt, right=3.5mm of top3] 
    (top4) {right hilar structures: There is no pulmonary edema. {[61, 66, 95, 110]}};

\draw[black, <-] (top1.south) to[out=-90, in=90] (T1);
\draw[green!70!black, <-] (top2.south) to[out=-90, in=90] (T2);
\draw[blue!70, <-] (top3.south) to[out=-90, in=90] (T3);
\draw[red!70!black, <-] (top4.south) to[out=-90, in=90] (T4);

\def\bottomoffset{3.5}
\def\bottomspread{2.5} 

\coordinate (B1) at ($ (decoder.south west)!0.15!(decoder.south east) $);
\coordinate (B2) at ($ (decoder.south west)!0.38!(decoder.south east) $);
\coordinate (B3) at ($ (decoder.south west)!0.62!(decoder.south east) $);
\coordinate (B4) at ($ (decoder.south west)!0.85!(decoder.south east) $);

\tikzset{
    botbox/.style={textbox, minimum width=2cm, align=center, inner sep=12pt}
}

\node[botbox, draw=black] (bot1) 
    at ($ (decoder.south) + (-3.5*\bottomspread, -\bottomoffset) $) 
    {Cap};
\node[botbox, draw=green!70!black, right=5mm of bot1] (bot2) {ARef};

\node[botbox, draw=blue!70, right=15mm of bot2] (bot3) {Gcap};

\node[botbox, draw=red!70!black, right=5mm of bot3] (bot4) {CARef};

\draw[black, <-] (B1) to[out=-90, in=90] (bot1.north);
\draw[green!70!black, <-] (B2) to[out=-90, in=90] (bot2.north);
\draw[blue!70, <-] (B3) to[out=-90, in=90] (bot3.north);
\draw[red!70!black, <-] (B4) to[out=-90, in=90] (bot4.north);

\end{tikzpicture}
\caption{\textbf{Overview of pretraining model architecture}. The model consists of a Siglip vision encoder and a transformer decoder. The vision encoder takes a chest X-ray image as input, produces visual tokens as cross-attention input to the transformer decoder. We adopt the following four tasks for pretraining: Cap, AREF, GCAP, and CAREF. }
\label{fig:locca-architecture}
\end{figure*}

\section{Methodology}

\subsection{Pretraining tasks}
We adopt a location-aware vision pretraining \cite{wan2024locca} within a unified multi-task generative learning framework \cite{beyer2023study}. Specifically, we incorporate localization information into the image captioning objective \cite{tschannen2023image} during pretraining using task-specific prefixes and a shared decoder to support unified multi-task learning, as illustrated in \autoref{fig:locca-architecture}.

Image captioning involves generating a sequence of tokens $y=[y_1,y_2,\ldots,y_n]$ that describe an image input $x$. To support multiple objectives within a unified generative framework, we prepend a task-specific prefix that conditions the model on the desired behavior. Across all location-aware tasks, each region is represented by a tuple consisting of a textual description and a bounding box defined by four coordinates $(x_{\min}, y_{\min}, x_{\max}, y_{\max})$.

We employ three such tasks. \textit{Automatic referring expressions (AREF)} requires the model to predict the bounding box coordinates corresponding to a given regional description.\textit{ Grounded captioning (GCAP)} tasks the model with generating a textual description for an image region specified by its bounding box coordinates. Finally, \textit{conditional automatic referring expressions (CAREF)} requires the model to jointly generate both the bounding box coordinates and the regional description given only an anatomical region name.
In all cases, the model is trained to generate bounding box coordinates and region descriptions sequentially, conditioned solely on the task prefix, rather than receiving either the coordinates or the descriptions as explicit inputs.

\subsection{Model details}
\subsubsection{Architecture}
We adopt a standard encoder–decoder architecture. The encoder consists of the SigLIP vision encoder \cite{zhai2023sigmoid}, followed by a two-layer multilayer perceptron (MLP) with GeLU activations that projects visual features into a shared multimodal embedding space. Text inputs are tokenized using a GPT‑2 tokenizer, mapped to embeddings through a linear projection, and augmented with learnable positional embeddings. The decoder follows a standard Transformer architecture \cite{vaswani2017attention} and integrates visual and textual information through cross-attention at every decoding layer.

During decoding, a causal attention mask ensures that each token is generated conditioned only on previously generated tokens. Followings prior work \cite{tschannen2023image, wan2024locca}, we additionally employ parallel token prediction for 25\% of the training samples in the standard image captioning task. In this setting, the model predicts all tokens independently and in parallel, relying solely on visual features. This strategy discourages over-reliance on preceding textual tokens and encourages stronger grounding of generated captions in visual content.

\subsubsection{Objective}
Model parameters are optimized by maximizing the conditional log-likelihood of the target output sequence,
\begin{equation}
    \sum_{i=1}^{|y|} \log P_\theta (y_i| y_{<i}, x)
\end{equation}
where each token $y_i$ is predicted conditioned on the image input $x$ and all previously generated tokens $y_{<i}$.
The model is trained to generate both captions and bounding box coordinates sequentially within a single output stream, rather than conditioning one prediction on the other (e.g., predicting captions given bounding boxes or vice versa). For a subset of training examples, we additionally employ parallel prediction, in which the model is trained to predict all output tokens simultaneously while preserving their correct order. This is implemented by masking all tokens in the input sequence and shifting the targets to align with the expected outputs.
The training loss is applied to all generated tokens except the task prefix. The loss on caption tokens encourages the model to accurately identify and describe the relevant image region, while the loss on bounding box coordinates refines the model’s ability to regress precise region locations corresponding to the generated descriptions.

\subsection{Datasets}
\subsubsection{MIMIC-CXR Database}
MIMIC‑CXR contains 377,110 chest X‑ray images from 65,379 patients, each study accompanied by a de‑identified radiology report \cite{johnson2024mimiccxr, goldberger2000physiobank, johnson2019mimiccxr}. The dataset includes longitudinal studies with multiple views per patient and follows official train, validation, and test splits.

Following prior work \cite{sellergren2025medgemmatechnicalreport, yang2025medical}, we restrict experiments to anterior–posterior and posterior–anterior views and use only the Findings and Impression sections, removing references to prior studies to avoid leakage. We use JPEG images from MIMIC‑CXR‑JPG \cite{johnson2024mimiccxrjpg} for efficiency, and employ this dataset for the image captioning task.

\subsubsection{Chest ImaGenome Dataset}
The Chest ImaGenome dataset \cite{wu2021chestimagenome} contains 242,072 frontal chest X‑ray images from MIMIC‑CXR, annotated with automatically constructed scene graphs. Each scene graph provides bounding boxes and descriptive phrases for a subset of 29 anatomical regions explicitly mentioned in the corresponding radiology report. The dataset follows the official MIMIC‑CXR training, validation, and test splits.

From these scene graphs, we extract region names, bounding box coordinates, and associated descriptions to construct training samples for our three location‑aware tasks. As illustrated in Figure~\ref{fig:locca-architecture}, AREF uses regional descriptions followed by bounding box coordinates, GCAP conditions on bounding box coordinates to generate descriptions, and CAREF conditions on anatomical region names and descriptions to generate bounding box coordinates.

\subsubsection{MIMIC-Diff-VQA Dataset}
MIMIC‑Diff‑VQA \cite{hu2023expert} contains 700,703 question–answer (QA) pairs constructed from 164,324 longitudinal chest X‑ray image pairs drawn from the MIMIC‑CXR dataset. The QA pairs are automatically generated from original radiology reports and span seven categories: difference, presence, abnormality, view, location, level, and type. In this work, we focus exclusively on the 164,324 QA pairs in the difference category, as they explicitly require comparative reasoning between a reference image and a follow‑up image. All experiments follow the official training, validation, and test splits provided with the dataset.

\subsection{Medical Difference VQA}
To address the medical difference visual question answering (Medical Diff-VQA) task, we trained the model on the MIMIC-Diff-VQA dataset. The model is comprised of a pre-trained vision encoder, a GPT-2 text decoder and a vision adapter (see \autoref{fig:med-diff-architecture}).

A shared vision encoder $E_{img}$ processes both the reference image ($i_{ref}\in \mathbb{R}^{H\times W}$) and the main image ($i_{main}\in \mathbb{R}^{H\times W}$), where $H=W=448$. Learnable temporal embeddings $t_{enc}^{ref}, t_{enc}^{main} \in \mathbb{R}^{n\times d_v}$ are added to the respective encoder outputs to encode temporal information:
\begin{equation}
\begin{split}
    v_{vis}^{ref} = E_{img}(i_{ref}) + t_{enc}^{ref} \in \mathbb{R}^{n\times d_v}\\
    v_{vis}^{main} = E_{img}(i_{main}) + t_{enc}^{main} \in \mathbb{R}^{n\times d_v}
\end{split}
\end{equation}
$n$ represents the number of visual tokens (flattened spatial features) and $d_v$ is the vision encoder's output dimension. A vision adapter, $P_{img}$, consisting of two linear layers with a GeLU activation is used to project the visual features into the multimodal embedding space. 
\begin{equation}
    \begin{split}
        v_{proj}^{ref} = P_{img}(v_{vis}^{ref}) \in \mathbb{R}^{n\times d_t}\\
        v_{proj}^{main} = P_{img}(v_{vis}^{main}) \in \mathbb{R}^{n\times d_t}
    \end{split}
\end{equation}
The question text $t$ is tokenized using the GPT-2 Tokenizer and mapped to embeddings $l_{txt} \in \mathbb{R}^{n \times d_t}$ via the model's native embedding layer. 
The input to the decoder is constructed by prepending the visual embeddings to the text embeddings as follows:
\begin{equation}
    Input_{seq} = [v_{proj}^{ref}; v_{proj}^{main}; l_{txt}]
\end{equation}
This results in a total sequence length of $2n+n_t$. 
Positional embeddings are added to the entire concatenated sequence.

A pre-trained GPT-2 \cite{radford2019language} medium decoder with 315 million parameters is used as the language decoder. The model is optimized using standard auto-regressive cross-entropy loss:
\begin{equation}
    L = -\sum_{j=1}^{|y|}log P_{\theta}(y_j|v_{proj}^{ref}, v_{proj}^{main}, l_{txt}, y_{<j})
\end{equation}
where $y$ represents the ground-truth answer tokens, and $P_\theta$ represents the trainable parameters. Crucially, loss is computed on the answer tokens only.

The vision encoder is frozen, while the vision adapter and the language decoder are finetuned. The model was trained using the AdamW optimizer (learning rate = $2\times10^{-4}$, weight decay = $1\times 10^{-2}$) with a label smoothing factor of 0.11. The model was trained for 10 epochs on an NVIDIA A100 GPU (80 GB VRAM), lasting 44 hours. The learning rate followed a linear warmup schedule for 10\% of the training steps, and a cosine decay schedule to zero.
\begin{table}[t]
\footnotesize
    \centering
    \caption{Results showing the performance of the pre-trained model on the grounded captioning task.}
    \label{tab:gcap-results}
    \begin{tabular}{lccc}
    \toprule
        BLEU-4& METEOR& ROUGE-L&CIDEr \\
         \midrule
         0.476 &0.501&0.593  &1.201\\
         \bottomrule
    \end{tabular}
\end{table}

\begin{table*}[t]
\footnotesize
  \centering
  \caption{Results showing the comparative performance of various vision encoder pretraining methods, each combined with a GPT-2 decoder, on the Medical Difference VQA task. The best results are in in \textbf{bold}. Under the same training settings, our method outperforms the baselines and the state of the art approaches.}
  \label{tab:overall_results}
  \begin{tabularx}{\textwidth}{@{} X ccccc @{}}
    \toprule
    \textbf{Method} & \textbf{BLEU-4 }& \textbf{METEOR} & \textbf{ROUGE-L} & \textbf{CIDEr} &\textbf{BertScore}\\
    \midrule
     Location Aware Pretraining (Ours) &\textbf{0.594} & \textbf{0.425} &\textbf{0.747}&\textbf{2.997} & \textbf{0.972}\\
     Global Contrastive Pretraining  & 0.291& 0.300 &0.407&0.550&0.850\\
     Regional Contrastive Pretraining &0.360 & 0.338 &0.465& 0.635 & 0.956\\
     BLIP-2 \cite{li2023blip} & 0.375  & 0.350& 0.545&0.801&0.960\\
     CapPa \cite{tschannen2023image}  & 0.350& 0.327&0.529&0.675&0.948\\
     RG-AG \cite{serra2025grounding} &0.551&0.384&0.668&2.198&0.965\\
     ReAL \cite{10.1007/978-3-031-72086-4_61} &0.530 &0.395 &0.736&2.409&0.968\\
         PLURAL \cite{cho2024pretraining} &0.520 &0.381&{0.653}&1.832&0.963\\
    \bottomrule
  \end{tabularx}
\end{table*}

\section{Experiments}
\subsection{Experimental setup}
We use an encoder–decoder model with 620 million parameters. The vision encoder follows the SigLIP architecture \cite{zhai2023sigmoid}, with 27 Transformer blocks, 12 attention heads, a hidden dimension of 1152, and a patch size of 14. The decoder comprises 12 Transformer blocks with 12 attention heads and a hidden dimension of 1024. All parameters are initialized randomly and trained from scratch.
The model is pretrained for five epochs on image–report pairs from MIMIC‑CXR and location‑aware samples from Chest ImaGenome. Each batch contains an equal number of samples from the four tasks. We optimize using AdamW ($\beta_1=0.9$, $\beta_2 0.999$) with 5k linear warm‑up steps followed by cosine annealing. The batch size is 32, with a learning rate of $10^{-4}$ and weight decay of $10^{-2}$. Images are resized to $448\times448$ without additional augmentation.

Text is tokenized using the GPT‑2 tokenizer \cite{radford2019language} (50,256 tokens). Bounding box coordinates are discretized into bins from 0 to 447 and represented as natural language integers, avoiding specialized location tokens. Training uses bfloat16 mixed precision and takes approximately 85 hours on a single NVIDIA A100 (80 GB).

\subsection{Vision Encoder Evaluation}
We evaluate the fine‑grained visual understanding of the pretrained vision encoder on a held‑out test set constructed from grounded captioning (GCAP) and conditional automatic referring expressions (CAREF) samples. We exclude automatic referring expressions (AREF) from evaluation due to their inherent ambiguity, as the same textual expression may validly correspond to multiple image regions, making ground‑truth comparison unreliable.

For the grounded captioning task, the model is prompted with a task prefix and bounding box coordinates, and the generated captions are evaluated using BLEU \cite{papineni2002bleu}, ROUGE‑L \cite{lin2004rouge}, METEOR \cite{banerjee2005meteor}, and CIDEr \cite{vedantam2015cider} (see Table~\ref{tab:gcap-results}). For the CAREF task, the model is prompted with the anatomical region name and evaluated on localization accuracy using the standard Acc@0.5 metric, achieving a score of 69.8\%.

\subsubsection{Baselines}
We evaluate our method against seven baselines on the difference category of the MIMIC‑Diff‑VQA dataset. For a controlled comparison, we re‑implement and train four baselines—Global Contrastive, Regional Contrastive, CapPa \cite{tschannen2023image}, and BLIP‑2 \cite{li2023blip}—using identical data splits and training settings. The Global Contrastive baseline uses a sigmoid‑based loss \cite{zhai2023sigmoid}, the Regional Contrastive model aligns image regions with captions, CapPa combines autoregressive and parallel prediction, and BLIP‑2 connects a frozen vision encoder to an LLM via a trainable Q‑Former.

We report results from prior work for the remaining baselines: PLURAL \cite{cho2024pretraining}, which models prior and follow‑up scans via multi‑stage pretraining; Pixel‑Difference (ReAL) \cite{10.1007/978-3-031-72086-4_61}, which operates on residual images; and RG‑AG \cite{serra2025grounding}, a two‑stage approach that uses generated radiology reports as intermediate reasoning context.

\begin{table*}[ht]
\footnotesize
\centering
\caption{Selected test cases to qualitatively compare the outputs of our approach with those of the baselines in Medical Diff VQA. \textbf{\textcolor{red}{Red}} indicates incorrect or missing predictions and \textbf{\textcolor{green}{green}} indicates correct predictions. The examples illustrate our model’s ability to identify differences between the two images and correctly attribute them to the appropriate image. However, there are instances where the model omits certain relevant features.}
\label{tab:qualitative_results}
\begin{tabular}{|l |l |l|} 
\hline
Case I & Case II & Case III \\
\hline
\qquad Reference  \qquad \qquad Main & \qquad Reference  \qquad \qquad Main & \qquad Reference \qquad \qquad Main\\
\includegraphics[width=0.3\textwidth, height=0.15\textwidth]{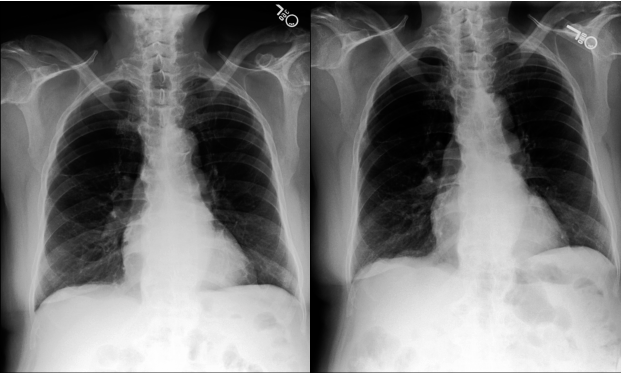} &
\includegraphics[width=0.3\textwidth, height=0.15\textwidth]{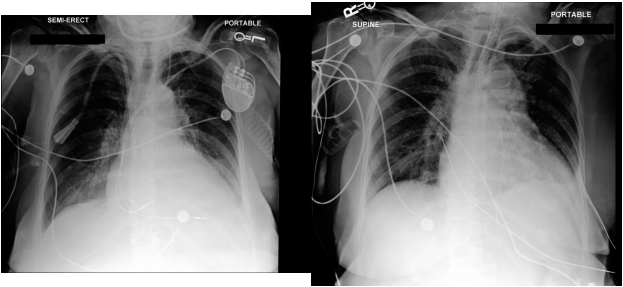} &
\includegraphics[width=0.3\textwidth, height=0.15\textwidth]{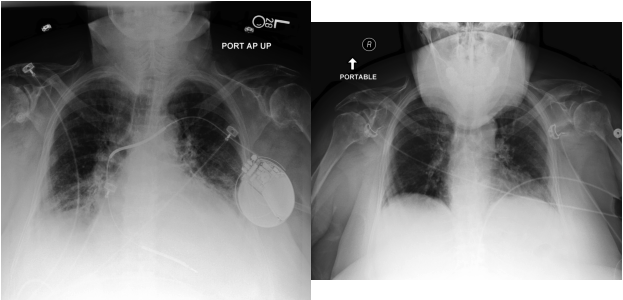} \\
\hline
\parbox[c]{0.25\textwidth}{\textbf{Q}: What has changed compared to the reference image?} &
\parbox[c]{0.25\textwidth}{\textbf{Q}: What has changed compared to the reference image?} &
\parbox[c]{0.25\textwidth}{\textbf{Q}: What has changed compared to the reference image?} \\
\hline
\parbox[c]{0.25\textwidth}{\textbf{GT}: the main image has additional findings of \textbf{\textcolor{green}{lung opacity}}, and \textbf{\textcolor{green}{atelectasis}} than the reference image. } &
\parbox[c]{0.25\textwidth}{\textbf{GT}: the main image has an additional finding of \textbf{\textcolor{green}{atelectasis}} than the reference image. } &
\parbox[c]{0.25\textwidth}{\textbf{GT}: the main image has additional findings of \textbf{\textcolor{red}{cardiomegaly, edema}}, \textbf{\textcolor{green}{atelectasis}}, and \textbf{\textcolor{green}{pleural effusion}} than the reference image. the main image is missing the findings of \textbf{\textcolor{green}{lung opacity}}, and \textbf{\textcolor{green}{pneumonia}} than the reference image.} \\
\hline
\parbox[c]{0.25\textwidth}{\textbf{Ours}: the main image has additional findings of \textbf{\textcolor{green}{lung opacity}}, and \textbf{\textcolor{green}{atelectasis}} than the reference image. } &
\parbox[c]{0.25\textwidth}{\textbf{Ours}: the main image has an additional finding of \textbf{\textcolor{green}{atelectasis}} than the reference image.} &
\parbox[c]{0.25\textwidth}{\textbf{Ours}: the main image has additional findings of \textbf{\textcolor{green}{atelectasis}}, and \textbf{\textcolor{green}{pleural effusion}} than the reference image. the main image is missing the findings of \textbf{\textcolor{green}{lung opacity}}, and \textbf{\textcolor{green}{pneumonia}} than the reference image.} \\
\hline
\parbox[c]{0.25\textwidth}{\textbf{ReAL}:  the main image has additional findings of \textbf{\textcolor{green}{lung opacity}}, and \textbf{\textcolor{green}{atelectasis}} than the reference image. } &
\parbox[c]{0.25\textwidth}{\textbf{ReAL}:  the main image has additional findings of \textbf{\textcolor{red}{pleural effusion}}, and \textbf{\textcolor{green}{atelectasis}} than the reference image.} &
\parbox[c]{0.25\textwidth}{\textbf{ReAL}:  the main image has additional findings of \textbf{\textcolor{green}{atelectasis}}, and \textbf{\textcolor{red}{lung opacity}} than the reference image.} \\
\hline
\end{tabular}
\end{table*}

\subsection{Quantitative Results}
In \autoref{tab:overall_results}, we compare different vision‑encoder pretraining strategies against state‑of‑the‑art baselines on the downstream medical difference VQA task using the MIMIC‑Diff‑VQA dataset. Model performance is evaluated using standard natural language generation metrics—BLEU, METEOR, ROUGE‑L, and CIDEr—which measure similarity between generated and reference answers based on n‑gram overlap and sequence alignment. To better capture semantic similarity beyond surface‑level lexical matching, we additionally report BERTScore \cite{zhang2019bertscore} computed with PubMedBERT (microsoft/BiomedNLP‑PubMedBERT‑base ‑uncased‑abstract‑fulltext).

The results show that our method consistently outperforms all baselines across evaluation metrics, with particularly large gains in BLEU‑4 and CIDEr. Relative to the strongest baseline, our model achieves a 7.8\% improvement in BLEU‑4 and a substantial 24.4\% gain in CIDEr, along with smaller but consistent improvements of 1.5\% in ROUGE‑L and 7.6\% in METEOR.

Beyond accuracy, our approach offers clear advantages in both training and inference efficiency. Unlike ReAL \cite{10.1007/978-3-031-72086-4_61}, it does not require explicit residual image computation or image registration, instead relying on learned visual representations to capture fine‑grained differences directly from paired inputs. In addition, our method outperforms PLURAL \cite{cho2024pretraining} while employing a significantly lighter training pipeline: whereas PLURAL relies on extensive multi‑stage pretraining across multiple domains, our approach achieves state‑of‑the‑art performance with a single, compute‑efficient pretraining stage.

\subsection{Qualitative Results}
\autoref{tab:qualitative_results} presents selected examples for a qualitative comparison between our model and ReAL \cite{10.1007/978-3-031-72086-4_61}. All responses are generated using beam decoding with a beam size of 10. In Case I, both models correctly identify the additional pathologies present in the main image. In Case II, ReAL fails to include pleural effusion among the additional findings in the main image. Case III represents a more challenging scenario, in which both models exhibit errors: ReAL incorrectly changes lung opacity from a missing finding to an additional finding in the main image and fails to mention pathologies present in the reference image but absent in the main image. Our model also makes errors in this case, omitting cardiomegaly and edema from the additional findings in the main image.
\begin{table*}[t]
\footnotesize
    \centering
    \caption{Results of the ablation study when we removed some of the location aware pretraining tasks. The best results are in \textbf{bold}.}
    \label{tab:ablation-study}
    \begin{tabular}{lcccccccccc}
    \toprule
    AREF& GCAP& CAREF& CAP&BLEU-4& METEOR & ROUGE-L & CIDEr \\
    \midrule
    \ding{51}&\ding{51}&\ding{51}&\ding{51}&\textbf{0.594}&\textbf{0.425}&\textbf{0.747}&\textbf{2.997}\\
    \ding{55}&\ding{51}&\ding{51}&\ding{51}&0.283&0.244&0.379&0.850\\
    \ding{51}& \ding{55} &\ding{51}&\ding{51}&0.347&0.318&0.527&0.945\\
    \ding{51}&\ding{51}& \ding{55}&\ding{51}&0.347& 0.316& 0.533&0.946\\
     \ding{55} &\ding{55}&\ding{55}&\ding{51}&0.350&0.317&0.529&0.950\\
    \bottomrule
    \end{tabular}
\end{table*}

\subsection{Ablation Study}
\subsubsection{Pretraining Tasks}
To assess the contribution of individual pretraining objectives, we conduct an ablation study focusing on the AREF, GCAP, CAREF, and captioning tasks. Specifically, we evaluate the effect of removing each task independently, as well as removing all location‑aware tasks altogether. All models are pretrained on the same dataset using an equal number of training samples at an image resolution of $448\times448$, and are subsequently fine‑tuned and evaluated on the medical difference VQA task. As shown in \autoref{tab:ablation-study}, removing all location‑aware tasks results in a substantial drop in performance across all evaluation metrics. Among the individual objectives, excluding AREF leads to the largest performance degradation, highlighting the importance of anatomical references for strengthening region‑level reasoning and fine‑grained visual understanding. Incorporating all location‑aware tasks yields the best overall performance, indicating that these objectives provide complementary benefits to the vision encoder’s

\subsubsection{Pretraining image resolution}
\autoref{tab:image-resolution-ablation} reports results for two pretraining input resolutions, 224 and 448. The models are trained under identical settings, differing only in input resolution, and are initialized from scratch at their respective resolutions. After pretraining, each model is fine-tuned and evaluated using the same resolution as during pretraining. We observe that the model trained at $448\times448$ consistently outperforms the model trained at $224\times224$ across all evaluation metrics.

\begin{table}[h]
\footnotesize
    \centering
    \caption{Ablation study of the impact of different image resolutions on the medical difference VQA. The best results are in \textbf{bold}.}
    \label{tab:image-resolution-ablation}
    \begin{tabular}{cccc}
    \toprule
         Image resolution& BLEU-4& METEOR& ROUGE\_L  \\
         \midrule
         $224\times224$&0.384&0.356&0.504\\
         $448\times448$&\textbf{0.594}&\textbf{0.425}&\textbf{0.747}\\
         \bottomrule
    \end{tabular}
\end{table}

\subsubsection{Parallel Prediction}
During pretraining, we randomly mask a subset of training examples and train the model to reconstruct the masked tokens in parallel while preserving their original order. To study the effect of this masking strategy on representation learning, we conduct an ablation in which the masking ratio is varied from 0\% to 50\%. All models are pretrained at a resolution of $448\times448$ and subsequently fine‑tuned on the medical difference VQA task at the same resolution. As shown in \autoref{tab:parallel-prediction}, applying parallel prediction to 25\% of the training examples yields the best downstream performance. In contrast, removing parallel prediction entirely results in a substantial performance drop, while increasing the masking ratio to 50\% similarly degrades performance. These results suggest that a moderate degree of parallel prediction provides an effective balance between sequential modeling and visually grounded learning.

\begin{table}[htb]
\footnotesize
    \centering
    \caption{Results showing the impact of the percentage of training samples that are masked during the vision encoder pretraining  on downstream performance of the vision encoder. The best results are in \textbf{bold}.}
    \label{tab:parallel-prediction}
    \begin{tabular}{c|ccc}
    \toprule
        Parallel (\%)&BLEU-4& METEOR& ROUGE-L \\
         \midrule
         0 &0.301&0.304  &0.448\\
         25&\textbf{0.594}& \textbf{0.425} &\textbf{0.747}\\
         50&0.358&0.339 & 0.484\\
         \bottomrule
    \end{tabular}
\end{table}

\section{Conclusion}
We introduce a location‑aware pretraining strategy for vision encoders tailored to the medical difference VQA task. By incorporating spatially grounded supervision, our approach enables the model to learn more fine‑grained representations of chest X‑ray images and improves its sensitivity to subtle inter‑image differences. As a result, the proposed method outperforms previous state‑of‑the‑art approaches.

Despite its effectiveness, our method relies on strong region‑level supervision, which is currently available only through the Chest ImaGenome dataset, limiting its immediate applicability to other imaging modalities. Additionally, although overall performance improves, the model occasionally omits or hallucinates certain pathologies when describing differences between image pairs. We hypothesize that scaling up the training data and increasing its diversity will help mitigate these failure modes and further enhance robustness.

\section*{Impact Statement}
This work aims to assist radiologists by automating the detection of temporal changes in medical imaging, potentially reducing diagnostic workload and improving the identification of disease progression. However, we acknowledge the risks associated with deploying VQA models in clinical settings.

First, like many generative models, our system may be prone to hallucinations or plausible but incorrect text generation; therefore, it is intended to function strictly as a human-in-the-loop assistive tool rather than a standalone diagnostic agent. Second, while our method improves robustness to acquisition differences, the model's performance is contingent on the diversity of the training data. Biases present in public chest X-ray datasets may propagate to the model's predictions. Future deployment requires rigorous validation across diverse patient populations to ensure equity and clinical safety.

\bibliography{example_paper}

@String(AAAI = {AAAI})

@InProceedings{10.1007/978-3-031-72086-4_61,
author="Lu, Zilin
and Xie, Yutong
and Zeng, Qingjie
and Lu, Mengkang
and Wu, Qi
and Xia, Yong",
editor="Linguraru, Marius George
and Dou, Qi
and Feragen, Aasa
and Giannarou, Stamatia
and Glocker, Ben
and Lekadir, Karim
and Schnabel, Julia A.",
title="Spot the Difference: Difference Visual Question Answering with Residual Alignment",
booktitle="Medical Image Computing and Computer Assisted Intervention -- MICCAI 2024",
year="2024",
publisher="Springer Nature Switzerland",
address="Cham",
pages="649--658",
isbn="978-3-031-72086-4"
}

@inproceedings{Hu_2023, series={KDD ’23},
   title={Expert Knowledge-Aware Image Difference Graph Representation Learning for Difference-Aware Medical Visual Question Answering},
   url={http://dx.doi.org/10.1145/3580305.3599819},
   DOI={10.1145/3580305.3599819},
   booktitle={Proceedings of the 29th ACM SIGKDD Conference on Knowledge Discovery and Data Mining},
   publisher={ACM},
   author={Hu, Xinyue and Gu, Lin and An, Qiyuan and Zhang, Mengliang and Liu, Liangchen and Kobayashi, Kazuma and Harada, Tatsuya and Summers, Ronald M. and Zhu, Yingying},
   year={2023},
   month=aug, pages={4156–4165},
   collection={KDD ’23} }

@misc{sellergren2025medgemmatechnicalreport,
      title={MedGemma Technical Report}, 
      author={Andrew Sellergren and Sahar Kazemzadeh and Tiam Jaroensri and Atilla Kiraly and Madeleine Traverse and Timo Kohlberger and Shawn Xu and Fayaz Jamil and Cían Hughes and Charles Lau and Justin Chen and Fereshteh Mahvar and Liron Yatziv and Tiffany Chen and Bram Sterling and Stefanie Anna Baby and Susanna Maria Baby and Jeremy Lai and Samuel Schmidgall and Lu Yang and Kejia Chen and Per Bjornsson and Shashir Reddy and Ryan Brush and Kenneth Philbrick and Mercy Asiedu and Ines Mezerreg and Howard Hu and Howard Yang and Richa Tiwari and Sunny Jansen and Preeti Singh and Yun Liu and Shekoofeh Azizi and Aishwarya Kamath and Johan Ferret and Shreya Pathak and Nino Vieillard and Ramona Merhej and Sarah Perrin and Tatiana Matejovicova and Alexandre Ramé and Morgane Riviere and Louis Rouillard and Thomas Mesnard and Geoffrey Cideron and Jean-bastien Grill and Sabela Ramos and Edouard Yvinec and Michelle Casbon and Elena Buchatskaya and Jean-Baptiste Alayrac and Dmitry Lepikhin and Vlad Feinberg and Sebastian Borgeaud and Alek Andreev and Cassidy Hardin and Robert Dadashi and Léonard Hussenot and Armand Joulin and Olivier Bachem and Yossi Matias and Katherine Chou and Avinatan Hassidim and Kavi Goel and Clement Farabet and Joelle Barral and Tris Warkentin and Jonathon Shlens and David Fleet and Victor Cotruta and Omar Sanseviero and Gus Martins and Phoebe Kirk and Anand Rao and Shravya Shetty and David F. Steiner and Can Kirmizibayrak and Rory Pilgrim and Daniel Golden and Lin Yang},
      year={2025},
      eprint={2507.05201},
      archivePrefix={arXiv},
      primaryClass={cs.AI},
      url={https://arxiv.org/abs/2507.05201}, 
}

@misc{li2023llavamedtraininglargelanguageandvision,
      title={LLaVA-Med: Training a Large Language-and-Vision Assistant for Biomedicine in One Day}, 
      author={Chunyuan Li and Cliff Wong and Sheng Zhang and Naoto Usuyama and Haotian Liu and Jianwei Yang and Tristan Naumann and Hoifung Poon and Jianfeng Gao},
      year={2023},
      eprint={2306.00890},
      archivePrefix={arXiv},
      primaryClass={cs.CV},
      url={https://arxiv.org/abs/2306.00890}, 
}

@misc{zhang2024pmcvqavisualinstructiontuning,
      title={PMC-VQA: Visual Instruction Tuning for Medical Visual Question Answering}, 
      author={Xiaoman Zhang and Chaoyi Wu and Ziheng Zhao and Weixiong Lin and Ya Zhang and Yanfeng Wang and Weidi Xie},
      year={2024},
      eprint={2305.10415},
      archivePrefix={arXiv},
      primaryClass={cs.CV},
      url={https://arxiv.org/abs/2305.10415}, 
}

@misc{zhang2025biomedclipmultimodalbiomedicalfoundation,
      title={BiomedCLIP: a multimodal biomedical foundation model pretrained from fifteen million scientific image-text pairs}, 
      author={Sheng Zhang and Yanbo Xu and Naoto Usuyama and Hanwen Xu and Jaspreet Bagga and Robert Tinn and Sam Preston and Rajesh Rao and Mu Wei and Naveen Valluri and Cliff Wong and Andrea Tupini and Yu Wang and Matt Mazzola and Swadheen Shukla and Lars Liden and Jianfeng Gao and Angela Crabtree and Brian Piening and Carlo Bifulco and Matthew P. Lungren and Tristan Naumann and Sheng Wang and Hoifung Poon},
      year={2025},
      eprint={2303.00915},
      archivePrefix={arXiv},
      primaryClass={cs.CV},
      url={https://arxiv.org/abs/2303.00915}, 
}

@inproceedings{radford2021learning,
  title={Learning transferable visual models from natural language supervision},
  author={Radford, Alec and Kim, Jong Wook and Hallacy, Chris and Ramesh, Aditya and Goh, Gabriel and Agarwal, Sandhini and Sastry, Girish and Askell, Amanda and Mishkin, Pamela and Clark, Jack and others},
  booktitle={International conference on machine learning},
  pages={8748--8763},
  year={2021},
  organization={PmLR}
}

@article{cho2024pretraining,
  title={Pretraining vision-language model for difference visual question answering in longitudinal chest x-rays},
  author={Cho, Yeongjae and Kim, Taehee and Shin, Heejun and Cho, Sungzoon and Shin, Dongmyung},
  journal={arXiv preprint arXiv:2402.08966},
  year={2024}
}

@inproceedings{bannur2023learning,
  title={Learning to exploit temporal structure for biomedical vision-language processing},
  author={Bannur, Shruthi and Hyland, Stephanie and Liu, Qianchu and Perez-Garcia, Fernando and Ilse, Maximilian and Castro, Daniel C and Boecking, Benedikt and Sharma, Harshita and Bouzid, Kenza and Thieme, Anja and others},
  booktitle={Proceedings of the IEEE/CVF Conference on Computer Vision and Pattern Recognition},
  pages={15016--15027},
  year={2023}
}

@inproceedings{yang2025medical,
  title={Medical Large Vision Language Models with Multi-Image Visual Ability},
  author={Yang, Xikai and Miao, Juzheng and Yuan, Yuchen and Wang, Jiaze and Dou, Qi and Li, Jinpeng and Heng, Pheng-Ann},
  booktitle={International Conference on Medical Image Computing and Computer-Assisted Intervention},
  pages={402--412},
  year={2025},
  organization={Springer}
}

@inproceedings{chen2025coca,
  title={CoCa-CXR: Contrastive Captioners Learn Strong Temporal Structures for Chest X-Ray Vision-Language Understanding},
  author={Chen, Yixiong and Xu, Shawn and Sellergren, Andrew and Matias, Yossi and Hassidim, Avinatan and Shetty, Shravya and Golden, Daniel and Yuille, Alan L and Yang, Lin},
  booktitle={International Conference on Medical Image Computing and Computer-Assisted Intervention},
  pages={78--88},
  year={2025},
  organization={Springer}
}

@article{he2022vlmae,
  title={VLMAE: Vision-language masked autoencoder},
  author={He, Sunan and Guo, Taian and Dai, Tao and Qiao, Ruizhi and Wu, Chen and Shu, Xiujun and Ren, Bo},
  journal={arXiv preprint arXiv:2208.09374},
  year={2022}
}

@inproceedings{huang2021gloria,
  title={Gloria: A multimodal global-local representation learning framework for label-efficient medical image recognition},
  author={Huang, Shih-Cheng and Shen, Liyue and Lungren, Matthew P and Yeung, Serena},
  booktitle={Proceedings of the IEEE/CVF international conference on computer vision},
  pages={3942--3951},
  year={2021}
}

@inproceedings{yao2022image,
  title={Image difference captioning with pre-training and contrastive learning},
  author={Yao, Linli and Wang, Weiying and Jin, Qin},
  booktitle={Proceedings of the AAAI Conference on Artificial Intelligence},
  volume={36},
  number={3},
  pages={3108--3116},
  year={2022}
}

@inproceedings{hu2023expert,
  title={Expert knowledge-aware image difference graph representation learning for difference-aware medical visual question answering},
  author={Hu, Xinyue and Gu, Lin and An, Qiyuan and Zhang, Mengliang and Liu, Liangchen and Kobayashi, Kazuma and Harada, Tatsuya and Summers, Ronald M and Zhu, Yingying},
  booktitle={Proceedings of the 29th ACM SIGKDD Conference on Knowledge Discovery and Data Mining},
  pages={4156--4165},
  year={2023}
}

@article{vaswani2017attention,
  title={Attention is all you need},
  author={Vaswani, Ashish and Shazeer, Noam and Parmar, Niki and Uszkoreit, Jakob and Jones, Llion and Gomez, Aidan N and Kaiser, {\L}ukasz and Polosukhin, Illia},
  journal={Advances in neural information processing systems},
  volume={30},
  year={2017}
}

@dataset{johnson2024mimiccxr,
  author       = {Johnson, Alistair and Pollard, Tom and Mark, Roger and Berkowitz, Seth and Horng, Steven},
  title        = {{MIMIC-CXR Database (version 2.1.0)}},
  year         = {2024},
  publisher    = {PhysioNet},
  note         = {RRID:SCR\_007345},
  doi          = {10.13026/4jqj-jw95},
  url          = {https://doi.org/10.13026/4jqj-jw95}
}

@article{goldberger2000physiobank,
  author       = {Goldberger, Ary L. and Amaral, Luis A. N. and Glass, Leon and Hausdorff, Jeffrey M. and Ivanov, Plamen Ch. and Mark, Roger G. and Mietus, Joseph E. and Moody, George B. and Peng, Chung-Kang and Stanley, H. Eugene},
  title        = {PhysioBank, PhysioToolkit, and PhysioNet: Components of a new research resource for complex physiologic signals},
  journal      = {Circulation},
  year         = {2000},
  volume       = {101},
  number       = {23},
  pages        = {e215--e220},
  note         = {RRID:SCR\_00734},
  url          = {https://www.ahajournals.org/doi/full/10.1161/01.CIR.101.23.e215}
}

@article{johnson2019mimiccxr,
  author       = {Johnson, Alistair E. W. and Pollard, Tom J. and Berkowitz, Seth J. and Greenbaum, Nathaniel R. and Lungren, Matthew P. and Deng, Chih-ying and Mark, Roger G. and Horng, Steven},
  title        = {{MIMIC-CXR, a de-identified publicly available database of chest radiographs with free-text reports}},
  journal      = {Scientific Data},
  year         = {2019},
  volume       = {6},
  number       = {317},
  doi          = {10.1038/s41597-019-0322-0},
  url          = {https://doi.org/10.1038/s41597-019-0322-0},
  publisher    = {Nature Publishing Group}
}

@dataset{johnson2024mimiccxrjpg,
  author       = {Johnson, Alistair and Lungren, Matthew and Peng, Yifan and Lu, Zhihao and Mark, Roger and Berkowitz, Seth and Horng, Steven},
  title        = {{MIMIC-CXR-JPG - chest radiographs with structured labels (version 2.1.0)}},
  year         = {2024},
  publisher    = {PhysioNet},
  note         = {RRID:SCR\_007345},
  doi          = {10.13026/jsn5-t979},
  url          = {https://doi.org/10.13026/jsn5-t979}
}

@dataset{wu2021chestimagenome,
  author       = {Wu, Jing and Agu, Nestor and Lourentzou, Ismini and Sharma, Abhishek and Paguio, Joshua and Yao, James S. and Dee, Erwin C. and Mitchell, William and Kashyap, Saurabh and Giovannini, Alessandro and Celi, Leo Anthony and Syeda-Mahmood, Tanveer and Moradi, Mehdi},
  title        = {{Chest ImaGenome Dataset (version 1.0.0)}},
  year         = {2021},
  publisher    = {PhysioNet},
  note         = {RRID:SCR\_007345},
  doi          = {10.13026/wv01-y230},
  url          = {https://doi.org/10.13026/wv01-y230}
}

@article{wan2024locca,
  title={Locca: Visual pretraining with location-aware captioners},
  author={Wan, Bo and Tschannen, Michael and Xian, Yongqin and Pavetic, Filip and Alabdulmohsin, Ibrahim M and Wang, Xiao and Susano Pinto, Andr{\'e} and Steiner, Andreas and Beyer, Lucas and Zhai, Xiaohua},
  journal={Advances in Neural Information Processing Systems},
  volume={37},
  pages={116355--116387},
  year={2024}
}

@article{beyer2023study,
  title={A study of autoregressive decoders for multi-tasking in computer vision},
  author={Beyer, Lucas and Wan, Bo and Madan, Gagan and Pavetic, Filip and Steiner, Andreas and Kolesnikov, Alexander and Pinto, Andr{\'e} Susano and Bugliarello, Emanuele and Wang, Xiao and Yu, Qihang and others},
  journal={arXiv preprint arXiv:2303.17376},
  year={2023}
}

@article{tschannen2023image,
  title={Image captioners are scalable vision learners too},
  author={Tschannen, Michael and Kumar, Manoj and Steiner, Andreas and Zhai, Xiaohua and Houlsby, Neil and Beyer, Lucas},
  journal={Advances in Neural Information Processing Systems},
  volume={36},
  pages={46830--46855},
  year={2023}
}

@inproceedings{zhai2023sigmoid,
  title={Sigmoid loss for language image pre-training},
  author={Zhai, Xiaohua and Mustafa, Basil and Kolesnikov, Alexander and Beyer, Lucas},
  booktitle={Proceedings of the IEEE/CVF international conference on computer vision},
  pages={11975--11986},
  year={2023}
}

@inproceedings{zhang2022glipv2,
  title={GLIPv2: unifying localization and VL understanding},
  author={Zhang, Haotian and Zhang, Pengchuan and Hu, Xiaowei and Chen, Yen-Chun and Li, Liunian Harold and Dai, Xiyang and Wang, Lijuan and Yuan, Lu and Hwang, Jenq-Neng and Gao, Jianfeng},
  booktitle={36th Conf. Neural Inf. Process. Syst. NeurIPS},
  year={2022}
}

@inproceedings{li2022grounded,
  title={Grounded language-image pre-training},
  author={Li, Liunian Harold and Zhang, Pengchuan and Zhang, Haotian and Yang, Jianwei and Li, Chunyuan and Zhong, Yiwu and Wang, Lijuan and Yuan, Lu and Zhang, Lei and Hwang, Jenq-Neng and others},
  booktitle={Proceedings of the IEEE/CVF conference on computer vision and pattern recognition},
  pages={10965--10975},
  year={2022}
}

@inproceedings{jia2021scaling,
  title={Scaling up visual and vision-language representation learning with noisy text supervision},
  author={Jia, Chao and Yang, Yinfei and Xia, Ye and Chen, Yi-Ting and Parekh, Zarana and Pham, Hieu and Le, Quoc and Sung, Yun-Hsuan and Li, Zhen and Duerig, Tom},
  booktitle={International conference on machine learning},
  pages={4904--4916},
  year={2021},
  organization={PMLR}
}

@inproceedings{tong2024eyes,
  title={Eyes wide shut? exploring the visual shortcomings of multimodal llms},
  author={Tong, Shengbang and Liu, Zhuang and Zhai, Yuexiang and Ma, Yi and LeCun, Yann and Xie, Saining},
  booktitle={Proceedings of the IEEE/CVF Conference on Computer Vision and Pattern Recognition},
  pages={9568--9578},
  year={2024}
}

@inproceedings{papineni2002bleu,
  title={Bleu: a method for automatic evaluation of machine translation},
  author={Papineni, Kishore and Roukos, Salim and Ward, Todd and Zhu, Wei-Jing},
  booktitle={Proceedings of the 40th annual meeting of the Association for Computational Linguistics},
  pages={311--318},
  year={2002}
}

@inproceedings{banerjee2005meteor,
  title={METEOR: An automatic metric for MT evaluation with improved correlation with human judgments},
  author={Banerjee, Satanjeev and Lavie, Alon},
  booktitle={Proceedings of the acl workshop on intrinsic and extrinsic evaluation measures for machine translation and/or summarization},
  pages={65--72},
  year={2005}
}

@inproceedings{lin2004rouge,
  title={Rouge: A package for automatic evaluation of summaries},
  author={Lin, Chin-Yew},
  booktitle={Text summarization branches out},
  pages={74--81},
  year={2004}
}

@article{radford2019language,
  title={Language models are unsupervised multitask learners},
  author={Radford, Alec and Wu, Jeffrey and Child, Rewon and Luan, David and Amodei, Dario and Sutskever, Ilya and others},
  journal={OpenAI blog},
  volume={1},
  number={8},
  pages={9},
  year={2019}
}

@inproceedings{fini2025multimodal,
  title={Multimodal autoregressive pre-training of large vision encoders},
  author={Fini, Enrico and Shukor, Mustafa and Li, Xiujun and Dufter, Philipp and Klein, Michal and Haldimann, David and Aitharaju, Sai and da Costa, Victor G Turrisi and B{\'e}thune, Louis and Gan, Zhe and others},
  booktitle={Proceedings of the Computer Vision and Pattern Recognition Conference},
  pages={9641--9654},
  year={2025}
}

@inproceedings{li2023blip,
  title={Blip-2: Bootstrapping language-image pre-training with frozen image encoders and large language models},
  author={Li, Junnan and Li, Dongxu and Savarese, Silvio and Hoi, Steven},
  booktitle={International conference on machine learning},
  pages={19730--19742},
  year={2023},
  organization={PMLR}
}

@inproceedings{vedantam2015cider,
  title={Cider: Consensus-based image description evaluation},
  author={Vedantam, Ramakrishna and Lawrence Zitnick, C and Parikh, Devi},
  booktitle={Proceedings of the IEEE conference on computer vision and pattern recognition},
  pages={4566--4575},
  year={2015}
}

@article{zhang2019bertscore,
  title={Bertscore: Evaluating text generation with bert},
  author={Zhang, Tianyi and Kishore, Varsha and Wu, Felix and Weinberger, Kilian Q and Artzi, Yoav},
  journal={arXiv preprint arXiv:1904.09675},
  year={2019}
}

@article{serra2025grounding,
  title={Grounding Chest X-Ray Visual Question Answering with Generated Radiology Reports},
  author={Serra, Francesco Dalla and Schrempf, Patrick and Wang, Chaoyang and Meng, Zaiqiao and Deligianni, Fani and O'Neil, Alison Q},
  journal={arXiv preprint arXiv:2505.16624},
  year={2025}
}
\bibliographystyle{icml2026}




\end{document}